\documentclass{article}

\usepackage{arxiv}

\usepackage[utf8]{inputenc} 
\usepackage[T1]{fontenc}    
\usepackage{hyperref}       
\usepackage{url}            
\usepackage{booktabs}       
\usepackage{amsfonts}       
\usepackage{nicefrac}       
\usepackage{microtype}      
\usepackage{lipsum}
\usepackage{graphicx}
\graphicspath{ {./images/} }

\usepackage{graphicx}
\usepackage{amsmath}
\usepackage{cite}
\usepackage{hyperref}
\usepackage{todonotes}
\usepackage{multirow}
\usepackage{makecell}
\usepackage{graphicx}

\usepackage{lipsum}

\usepackage{varwidth} 
\usepackage{setspace}

\counterwithout{table}{section}  
\usepackage{chngcntr}

\title{Efficient extraction of medication information from clinical notes: an evaluation in two languages
}

\author{
Thibaut FABACHER, MD \\
University hospital of Strasbourg \\
ICube Laboratory, Strasbourg, France \\
Inria, Inserm, Université Paris Cité, U1346 HeKA, Paris, France\\
\And
Erik-André SAULEAU, PhD \\
University hospital of Strasbourg \\
ICube Laboratory, Strasbourg, France\\
\And
Emmanuelle ARCAY\\
University hospital of Strasbourg \\
\And
Bineta FAYE\\
University hospital of Strasbourg \\
\And
Maxime ALTER\\
University hospital of Strasbourg \\
\And
Archia CHAHARD \\
University hospital of Strasbourg\\
\And
Nathan MIRAILLET \\
University hospital of Strasbourg\\
\And
Adrien COULET, PhD\\
Inria, Inserm, Université Paris Cité, U1346 HeKA, Paris, France \
\And
Aurélie NÉVÉOL, PhD\\
Université Paris-Saclay, CNRS, LISN, Orsay, France \
}

\date{}
\begin{document}
\maketitle
\begin{abstract}
\textbf{Objective}: To evaluate the accuracy, computational cost and portability of a new Natural Language Processing (NLP) method
for extracting medication information from clinical narratives.
\textbf{Materials and Methods:} We propose an original transformer-based architecture for the extraction of entities and their relations
pertaining to patients' medication regimen. First, we used this approach to train and evaluate a model on French clinical notes,
using a newly annotated corpus from Hôpitaux Universitaires de Strasbourg. Second, the portability of the approach was as-
sessed by conducting an evaluation on clinical documents in English from the 2018 n2c2 shared task. Information extraction
accuracy and computational cost were assessed by comparison with an available method using transformers.
\textbf{Results:} The proposed architecture achieves on the task of relation extraction itself performance that are competitive with the
state-of-the-art on both French and English (F-measures 0.82 and 0.96 vs 0.81 and 0.95), but reduce the computational cost
by 10. End-to-end (Named Entity recognition and Relation Extraction) F1 performance is 0.69 and 0.82 for French and English
corpus.
\textbf{Discussion:} While an existing system developed for English notes was deployed in a French hospital setting with reasonable
effort, we found that an alternative architecture offered end-to-end drug information extraction with comparable extraction per-
formance and lower computational impact for both French and English clinical text processing, respectively.
\textbf{Conclusion:} The proposed architecture can be used to extract medication information from clinical text with high performance
and low computational cost and consequently suits with usually limited hospital IT resources
\end{abstract}

\section*{Introduction}

Medication information is crucial to provide patients with medical care tailored to their condition. 
Understanding detailed therapeutic regimens is highly valuable, especially for patients with chronic diseases who may receive care across multiple facilities, including hospitals. Accurate representation of these regimens supports improved clinical research and facilitates better management of complex, long-term treatments. 

Electronic health records (EHRs) are a major source of clinical data, providing comprehensive information for patients treated at a hospital. EHRs include both structured and unstructured data sources. While structured data is often the focus of clinical research, unstructured data represents an untapped resource that can yield significant clinical insights~\cite{demner2009can,escudie2017novel}. Among the various aspects of unstructured clinical data,  medication-related information is particularly  valuable.

Medication information extraction from EHRs involves identifying specific drugs, with their attributes such as dosages, frequencies, routes, duration and their contextualization  within a patient’s clinical timeline. Elements of contextualization can be dosage adjustments, therapy modifications, negation, proposition of treatment, etc. The international biomedical Natural Language Processing (NLP) community has been addressing this issue over the past decade. Despite notable success for English, this task remains challenging due to the variability of language in clinical texts and the complexity to account for heterogenous contextual factors~\cite{uzuner2010extracting,MODI2024104603}.
Existing approaches typically focus on either drugs and their attributes or the contextual factors associated with them. Few methods capture the full scope of prescriptions, which involve specific combinations of drugs and their attributes tied to specific points in a patient's timeline. This gap leads to imprecise or incomplete therapeutic schemes, limiting their utility in real-world clinical applications.

Recent advances in NLP, particularly transformer-based architectures, have significantly enhanced performance in tasks such as Named Entity Recognition (NER) and Relation Extraction (RE) \cite{FRAILENAVARRO2023105122,Landolsi2023}. Models like Bidirectional Encoder Representations from Transformers (BERT) \cite{BERT} and Generative Pre-trained Transformers (GPT) \cite{GPT}, achieve this by leveraging contextual word embeddings. 

One major issue is computational complexity. These models can be extremely large, requiring substantial memory and processing powerfor either training or fine-tuning
on domain-specific datasets. Additionally, models like BERT and GPT are often used as input of classifiers for relationships between entity pairs, where each pair is processed individually. This leads to high computational costs. Moreover, models like GPT, are particularly resource-intensive, making them impractical for low-resource environments ~\cite{gema-etal-2024-edinburgh}.

Another challenge is data privacy. Clinical data is highly sensitive and subject to strict regulations, which often prohibit the use of external APIs or cloud-based systems. This necessitates  the local deployment of resource-efficient models for RE.

Adapting models to real-world clinical datasets and settings
presents additional challenges. Clinical notes are often noisy, inconsistently formatted, and highly variable in linguistic structure, which complicates model generalization~\cite{Carrell2017}. Furthermore, studies have shown that while models trained on synthetic or curated datasets perform well in controlled environments, their performance often degrades when applied to real-world corpora~\cite{miller-etal-2021-domain}. The disparity between training data and real-world text underscores the need for robust methods that can handle the variability and complexity of actual clinical data within a constrained computational environment.

\section*{Objectives}
The primary objective of this study is to propose a method for extracting drug-related information from French patient medical records. This method is designed to be computationally efficient, 
as it will be deployed  in
an environment with low computational resources.
A secondary objective is to validate the proposed method on an existing English dataset from the n2c2 challenge, 
to assess portability
across languages and 
provide a
comparison with state-of-the-art approaches.

\subsubsection*{Contributions}
This study makes two key contributions. First, we introduce an innovative architecture for RE that significantly reduces computational cost, while achieving state-of-the-art performance. This approach addresses the limitations of current models, particularly in low-resource and high-privacy clinical environments. It was initially designed for French clinical documents but also achieves state-of-the-art performance on English clinical text. Transitioning between languages, requires only a change in the transformer model, highlighting the adaptability of our approach.
Second, we redefine how relationships between drugs and associated entities should be annotated in free text, to offer a more precise and comprehensive representation of therapeutic regimens. 
\subsection*{Research Questions}
Our study addresses the following questions: 
    \begin{itemize}
    \item Can we reduce the computational burden of the RE task without compromising accuracy?
    \item How effectively can we capture nuanced relationships between drugs and their attributes, particularly in scenarios with complex temporal dependencies ?
    \item Can we generalize to languages other than English, particularly French clinical texts, while retaining high performance?

    \end{itemize}

\section*{Materials and Methods}

\subsection*{Representation of drugs and their attributes}

Current representations often treat drug names and drug attributes as independent entities, which fails to support the accurate description of therapeutic regimens over time. For example, consider the following prescription from the French corpus, translated into English for simplicity: 
\begin{verbatim}treatment with tocilizumab IV every 
4 weeks from July to October, then every 
2 weeks until December \end{verbatim}
The drug (tocilizumab) is prescribed with a different frequency (every 2 weeks, every 4 weeks)  in two periods (July to October, October to December), so that the same drug will need to be linked two distinct pairs of prescription frequency/period. Drug representations need to account for therapy adjustments over time. Existing methods tend to link every attributes to the associated drug, making it difficult to capture such nuances, particularly common in chronic disease management.

\begin{figure}
    \centering
    \includegraphics[width=1\linewidth]{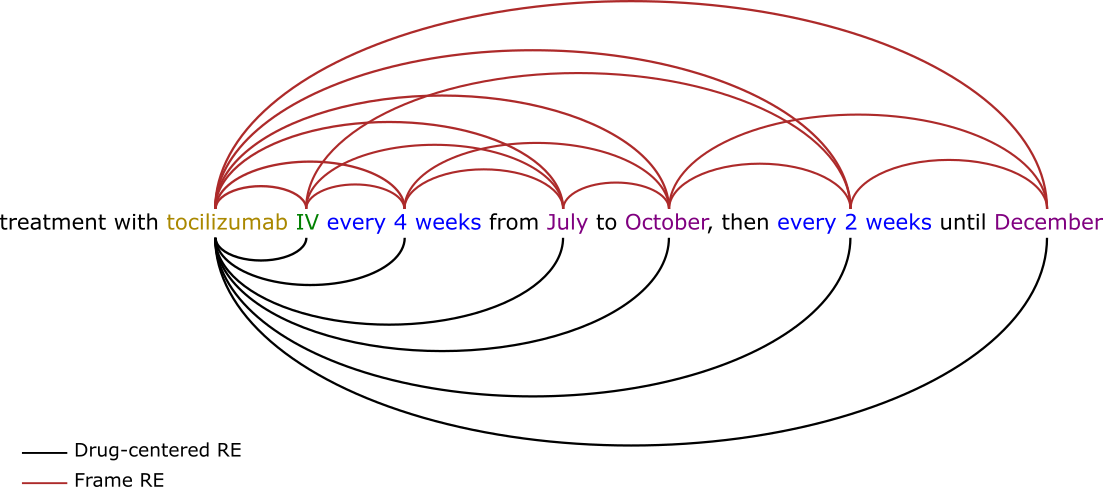}
    \caption{Comparative illustration of two concurrent types of annotations for medication relations : drug-centered vs. frame relations. In the case of drug-centered relations (bottom annotations), each relation starts from the drug name. In the case of the frames, each frame captures a specific drug, with specific attributes such as strength, dose, route of administration, and prescription duration. Relationships are visualized as links between attributes within a frame, providing a structured view of drug-attribute relations. In this example, there are two frames corresponding to two distinct pairs of prescription frequency/period}
    \label{fig:same_frame_relation}
\end{figure}

To improve the representation of relationships between drugs and their attributes, we propose to use the concept of frames\cite{baker-etal-1998-berkeley-framenet}. A frame corresponds to a specific drug with associated properties, including strength, dose, route of administration, and prescription duration. This richer representation is shown in Figure \ref{fig:same_frame_relation}. This framing mechanism was integral to our annotation process and provided a structured approach for capturing complex drug-attribute interactions.

\subsection*{Clinical corpora}

To evaluate the proposed model, we used both an English and a French corpora. The English corpus, n2c2\cite{Henry2020-wp}, was employed to benchmark our approach against state-of-the-art methods, while the French corpus was curated to test the model's performance on domain-specific clinical text and test our proposed data representation scheme.

\begin{figure}
    \centering
    \includegraphics[width=1\linewidth]{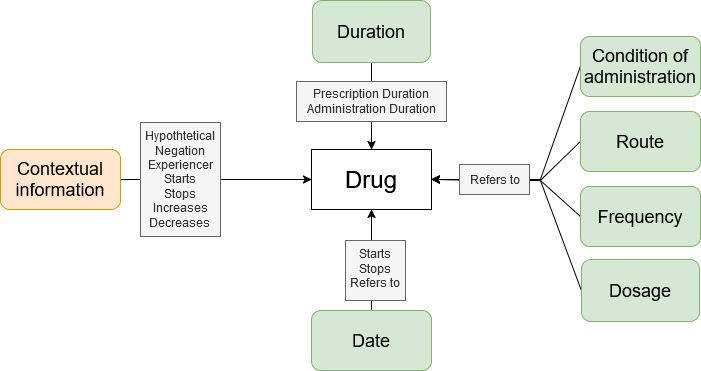}
    \caption{Representation of Treatment Regimen: This figure illustrates the representation of a treatment regimen, defined as a specific drug and its associated attributes. Some attributes, such as Dosage and Frequency, have non-specific relation types to the drug, while others, like dates, can trigger more specific relation types (starting or stopping a medication). Contextual information enables the representation of these specific relation types without requiring a distinct entity type. A frame is triggered by a drug and includes a specific set of attributes. Among these attributes, only the drug itself is mandatory to create a frame.}
    \label{fig:treatment_regimen}
\end{figure}

\subsubsection*{French Corpus: Corp-HUS }
The French clinical corpus, named Corp-HUS, was collected from Strasbourg University Hospital. This study is listed on hospital study register and follows the hospital clinical study protocol. It consists of 715 documents extracted from EHRs of patients diagnosed with probable rheumatoid arthritis (RA), including discharge summaries, paramedical and admission notes. These documents were annotated following a detailed annotation guide as part of a larger project aimed at extracting therapeutic schemas for RA patients\footnote{\url{https://github.com/DrFabach/guide_annotation_medicaments}}. 
Those annotations include the following entities: Drug, Class of Drug, Date, Relative Date, Dosage, Frequency, Route, Duration, and Context; and relations : Refer\_to, Start, Stop, Augmentation, Diminution, Contraindication, Negation, Hypothetical, Experiencer, Administration Duration, Prescription Duration, Coreference, and Discontinue Entity. The Refer\_to relation establishes a generic link between a drug name and its attributes like Dosage, Frequency, Route, or Date (Figure \ref{fig:treatment_regimen}).

The corpus is divided into two subsets: a train set of 615 documents and a test set of 100 documents. It contains  7,581 entities and 4,293 relations in the train set, and 1,726 entities and 904 relations in the test set (Table \ref{tab:table_relation}). Notably, one relation type, Refer\_to, constitutes the majority of relations, accounting for over half of the annotated links. In training set, 2,427 entities are identified as drugs, of which 1,851 frames can be constructed, with 3.9\% classified as multi-frame drugs (drugs involved in several frames, as in Figure \ref{fig:same_frame_relation}).

Due to data privacy policies, the French corpus cannot currently be shared. However, we are committed to anonymizing our corpus to enable data sharing with other academics.

In our dataset, inter-annotator agreement was measured to evaluate the consistency of manual annotations using F1 score for relations. The mean micro-average F1-score between human annotators was 0.58 (sd: 0.28). When comparing annotations between human annotators and the gold standard, the F1-score increased to 0.68 (sd: 0.1). For automated pre-annotation methods, rule-based dictionary approaches yielded an F1-score of 0.37, while the use of a preliminary trained algorithm improved the F1-score to 0.52.

\subsubsection*{English Corpus: the 2018 n2c2 }
The  corpus used in the 2018 n2c2 Challenge on Adverse Drug Events and Medication Extraction \cite{Henry2020-wp} comprises tasks for NER and RE. The concept extraction task involves identifying the following entities: Drug, Strength, Form, Dosage, Frequency, Route, Duration, Reason, and ADE (Adverse Drug Event). The relation classification task defines links between entities, including Strength-Drug, Form-Drug, Dosage-Drug, Frequency-Drug, Route-Drug, Duration-Drug, Reason-Drug, and ADE-Drug.

The corpus is divided into two subsets: a train and a test set with 303 and 202 discharge notes, respectively. The n2c2 corpus \cite{Henry2020-wp} includes 50,951 entities and 36,384 relations in the train, and 32,918 entities and 23,462 relations in the test set (Table \ref{tab:table_relation}).

\begin{table*}
\centering

\caption{Distribution of relations in the two corpora, n(\%). \dag \textit{Refer\_to} relation in Corp-HUS. \ddag \textit{Duration-Drug} relation in n2c2 corpus }
\label{tab:table_relation}
\begin{tabular}{lrrrr} 

                      & \multicolumn{2}{c}{\textbf{Corp-HUS}}                                 & \multicolumn{2}{c}{\textbf{n2c2}}                                   \\\cmidrule(l){2-3}\cmidrule(l){4-5}
                      & Train (\%)                    & Test (\%)                    & Train (\%)                  & Test (\%)                    \\ 

\hline
Total doc             & \multicolumn{1}{r}{615}       & \multicolumn{1}{r}{100}      & \multicolumn{1}{r}{303}     & \multicolumn{1}{r}{202}      \\
\cmidrule(l){1-5}
Total entities        & \multicolumn{1}{r}{7,581}     & \multicolumn{1}{r}{1,726}    & \multicolumn{1}{r}{50,951}  & \multicolumn{1}{r}{36,384}   \\
\cmidrule(l){1-5}
Total relations       & \multicolumn{1}{r}{4,293}     & \multicolumn{1}{r}{904}      & \multicolumn{1}{r}{32,918}  & \multicolumn{1}{r}{23,462}   \\
\hline\hline
Strength→Drug         & \multirow{5}[3]{*}{\shortstack[c]{2515 (58.58)\\\dag}} & \multirow{5}[3]{*}{\shortstack[c]{488 (53.98)\\\dag}} & 6,702 (18.42)               & 4,244 (18.09)                \\
\cmidrule(l){1-1}\cmidrule(l){4-5}
Form→Drug             &                               &                              & 6,654 (18.29)               & 4,374 (18.64)                \\
\cmidrule(l){1-1}\cmidrule(l){4-5}
Dosage→Drug           &                               &                              & 4,225 (11.61)               & 2,695 (11.49)                \\
\cmidrule(l){1-1}\cmidrule(l){4-5}
Frequency→Drug        &                               &                              & 6,310 (17.34)               & 4,034 (17.19)                \\
\cmidrule(l){1-1}\cmidrule(l){4-5}
Route→Drug            &                               &                              & 5,538 (15.22)               & 3,546 (15.11)                \\
\cmidrule(l){1-5}
Start                 & 531 (12.37)                   & 121 (13.38)                  & -                           & -                            \\
\cmidrule(l){1-5}
Stop                  & 307 (7.15)                    & 70 (7.74)                    & -                           & -                            \\
\cmidrule(l){1-5}
Ongoing               & 420 (9.78)                    & 123 (13.61)                  & -                           & -                            \\
\cmidrule(l){1-5}
Duration prescription & 51 (1.9)                      & 21 (2.32)                    & \multirow{2}[3]{*}{\shortstack[c]{643 (1.77)\\\ddag}} & \multirow{2}[3]{*}{\shortstack[c]{426 (1.82)\\\ddag}}  \\
\cmidrule(l){1-3}
Administration time   & 12 (0.28)                     & -                            &                             &                              \\

\cmidrule(l){1-5}
Increase              & 44 (1.02)                     & 10 (1.11)                    & -                           & -                            \\
\cmidrule(l){1-5}
Decrease              & 33 (0.77)                     & 10 (1.11)                    & -                           & -                            \\
\cmidrule(l){1-5}
Negation              & 54 (1.26)                     & 12 (1.33)                    & -                           & -                            \\
\cmidrule(l){1-5}
Contraindicated       & 46 (1.07)                     & 5 (0.55)                     & -                           & -                            \\
\cmidrule(l){1-5}
Hypothetical          & 48 (1.12)                     & 9 (1)                        & -                           & -                            \\
\cmidrule(l){1-5}
Experiencer           & 1 (0.02)                      & -                            & -                           & -                            \\
\cmidrule(l){1-5}
Coref                 & 152 (3.54)                    & 34 (3.76)                    & -                           & -                            \\
\cmidrule(l){1-5}
Discontinue           & 34 (0.79)                     & 1 (0.11)                     & -                           & -                            \\
\cmidrule(l){1-5}
Reason→Drug & -                           & -           & 5,169 (14.21)                 & 3,410 (14.53)                                           \\
\cmidrule(l){1-5}
ADE→Drug & -                           & -              & 1,107 (3.04)                  & 733 (3.12)                                              \\
\hline
\end{tabular}
\end{table*}

\subsection*{Relation extraction with NLP}

\subsubsection*{Task description}

The task of extracting drugs and their attributes from unstructured clinical text involves two key components: NER and RE \cite{Landolsi2023}. Although some studies combine these two steps into a single one \cite{survey,comprehensive_survey}, separation can optimize performance \cite{frustratingly}, particularly when using distinct token representations for each task, which can improve overall results.
Given the maturity of NER solutions, the focus of this work is on improving the RE component.

\subsubsection*{Comparison to a transformer-based relation classification method} 

For RE, a common strategy involves tagging the two entities within a sentence and using a transformer augmented with an additional classification layer to determine the existence and type of relationship\cite{frustratingly,zheng-etal-2017-joint,mtumbuka-schockaert-2024-entity}. However, these methods face a significant challenge in terms of computation time. Each pair of entities must be presented to the model individually, drastically increasing training time. 
We compared our proposed architecture with usual models, such as the transformer-based approaches described in \cite{frustratingly,Mahendran2021-dj,Wei2020-ab,yang2021clinicalrelationextractionusing}. Those classify relationships by marking entities within sentences and performing classification on the transformer output representations. A notable comparison system selected, detailed in \cite{yang2021clinicalrelationextractionusing}, duplicates sentences containing entities for classification purposes. This method was selected because it achieves good state-of-the-art results on n2c2 corpus, and implementation code was publicly available as a Python package \footnote{\url{https://github.com/uf-hobi-informatics-lab/ClinicalTransformerRelationExtraction}}.

\subsubsection*{Proposed Architecture}

\begin{figure*}

    \centering
    \includegraphics[width=1\linewidth]{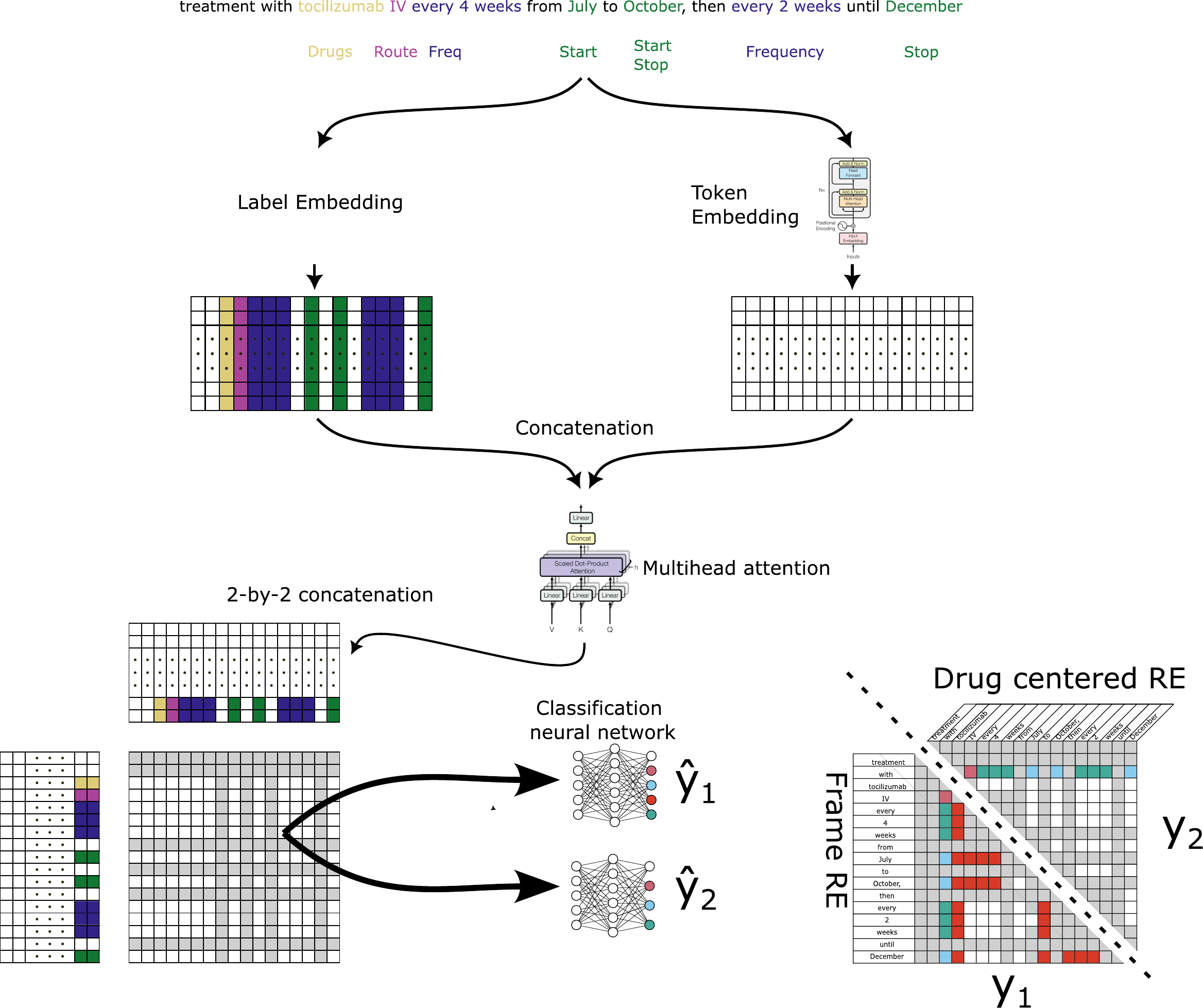}
    \caption{Overview of Our Model Architecture: The model uses a Transformer-based architecture to classify all relationships simultaneously. Input sentences are tokenized, and contextual embeddings are generated via a Transformer layer, while a dedicated embedding layer is simultaneously applied to token labels. These two representations are concatenated to form a unified token representation, which is passed through a multi-head self-attention layer to capture token dependencies. Subsequently, pairwise combinations of tokens are combined with their relative position embeddings and processed through a deep neural network consisting of a dense layer followed by a task-specific classification layer. The classification layer's output nodes correspond to the number of relation types being predicted. The model is trained using a masked cross-entropy loss applied exclusively to label token pairs, ensuring efficient optimization for the complexity of relational tasks.
The matrix represents the two different approaches to representing a drug and its attributes.}
    \label{fig:our_model}

\end{figure*}

Our model depicted Figure \ref{fig:our_model} is designed to classify all relationships simultaneously, similar to Zhang et al. \cite{Zhang}. It employs a standard transformer architecture to tokenize input sentences and generate contextual embeddings for each token, embedding size depending of the pretrained transformer model used. Additionally, we add a dedicated embedding layer for token labels.

These embeddings are concatenated to form a unified token representation, which is in turn passed through a multi-head self-attention layer, leveraging the attention mechanism to capture dependencies. To extract relationships, token pairs are represented by concatenating their embeddings and incorporating relative position embeddings to encode their positional relationship. These token-pair representations are then processed through one dense layer followed by a classification layer.

During training, we optimize computation by applying a mask to the loss function, restricting its application to labeled token pairs, i.e. token associated with an entity label. The final classification layer is a multiclass classifier, trained using the Adam optimizer with a linear scheduler incorporating a warmup phase. The loss function employed is cross-entropy.

We leveraged pretrained transformer models fine-tuned on clinical datasets to produce domain-specific word representations. For the English dataset, we experimented with ClinicalBERT\cite{clinicalbert} and BioBERT\cite{biobert}, while for the French corpus we used CamemBERT-BIO\cite{camembert_bio}. Default tokenizers were used for each transformer model.

\subsubsection*{Preprocessing}

The French corpus is directly extracted from EHRs where newlines are frequently used to indicate the end of sentences. To address this, we incorporated the token \texttt{\textbackslash n} into the tokenizer to handle newline characters effectively and preserve the structure of the text. 
 
Unlike prior transformer-based models, our approach eliminates the need for extensive preprocessing. The only required inputs are the text and entity labels for RE tasks. To optimize computational efficiency, we constrained the input text length using a sliding window of $n$ characters. Segments containing fewer than two labeled entities were excluded from train and test datasets. Beyond this, no additional  preprocessing was required.

\subsubsection*{Hyperparameter selection}
We selected the best hyperparameters through a grid search cross-validation on the train set. The optimized parameters included the learning rate, the dimensionality of label embeddings, the number of heads in the multi-head attention layer, the dimensionality of positional embeddings, and the number of nodes in the hidden layer of the final classifier.

\subsubsection*{End-to-end}

For end-to-end drug and attribute RE, we use the NLStruct library~\cite{wajsburt} for NER and apply our proposed method for RE sequentially.

\subsection*{Evaluation Metrics}

Performance comparisons between algorithms are conducted using precision, recall and F1 scores as evaluation metrics for each category. Among these, the micro-average F1 score serves as the principal value for assessing and comparing architectures, as it emphasizes performance across all instances equally. 
For the n2c2 dataset, we used the evaluation script  provided by the 2018 n2c2 Challenge and compared performance using the micro-average F1-score, precision, and recall. For French corpus, we adapt the n2c2 evaluation script to our task.

To assess computational efficiency, we tracked the training times using Python's datetime library and TensorBoard\cite{tensorboard}.

\subsubsection*{Sensitivity analysis}
To compare performance between the Corp-HUS and n2c2 corpora, we randomly reduced the n2c2 training set to match our Corp-HUS size by selecting five separate subsets of 25 documents. This ensured a comparable number of entities between the n2c2 subsets and the French corpus.

\subsection*{Work environment and implementation}

Our architecture will be made available in the medkit library\cite{medkit}.
The experiments were conducted on a Windows system equipped with an NVIDIA RTX 3070 GPU, an Intel(R) Xeon(R) E3-1245 v5 CPU, 64 GB of RAM, and Python 3.9. The main Python libraries used in this study were medkit \cite{medkit}, PyTorch\cite{pytorch}, Transformers\cite{transformers}, Pandas\cite{pandas}, NumPy\cite{numpy} and SeqEval\cite{seqeval}. Ecological impact was assessed using the Green Algorithms\cite{Green_Algorithms}.

\section*{Results}

\subsection*{Performance}
The grid search process yielded consistent optimal hyperparameters for both the French and English corpora. The best-performing configuration included a batch size of 10, 4 attention heads for the additional multi-head attention layer, an embedding size of 32 nodes for entity labels, 75 nodes for positional embeddings, 256 nodes for the hidden layer, and a learning rate of 0.0001.
\subsubsection*{Performance on the English Corpus}
On the n2c2 corpus, our proposed model achieved an F1 score of 0.955 (table: \ref{tab:f1_score}) which is comparable to the performance of Yang et al. 2021\cite{yang2021clinicalrelationextractionusing} algorithm (0.949). Notably, our method demonstrated a significant advantage in computational efficiency, being approximately 23 times faster during training.
The choice of Transformer models (e.g., ClinicalBERT or BioBERT) did not influence the F1 score, suggesting that the model performance is robust to the selection of pretrained Transformer variants within this domain.

\subsubsection*{Performance on the French Corpus}
On the French corpus, our model achieved an F1 score of 0.807 using a sliding window size of 300 characters, which is comparable to Yang et al.'s algorithm (0.809)~\cite{yang2021clinicalrelationextractionusing}. However, our approach demonstrated improved precision (0.834 vs. 0.786) at the cost of reduced recall (0.782 vs. 0.834). 
Increasing the sliding window size from 200 characters to 300 enhanced the recall by three points, reflecting the model’s improved ability to capture long-range dependencies. Additionally, the integration of frame information led to an improvement in both precision and the global F1 score (0.821 vs. 0.807), emphasizing the potential significance of this additional contextual information for accurate RE.
Despite achieving similar overall performance metrics, our algorithm demonstrated a significant computational advantage, training approximately 10 times faster than the baseline Transformer architecture.

\begin{table*}
\centering

\caption{Performance comparison on French and English corpora, micro-average scores (mean (sd) over $5$ runs). sd was calculated only on French corpus, to limit computational time. Models used: CamemBERT-BIO (French) and ClinicalBERT (English). Mean training time in minutes. 
* indicates statistical difference with Yang et al. 2021. \dag Reference Model from Literature for Comparison \ddag 3 best teams n2c2 challenge\cite{Henry2020-wp}}
\label{tab:f1_score}
\resizebox{\textwidth}{!}{
\renewcommand{\arraystretch}{1.2}  
\begin{tabular}{clllllr}

& Corpus & Model (sliding windows of $n$ characters) & Precision & Recall & F1 & Time \\
\cline{1-7}
\multirow{11}{*}{\rotatebox{90}{RE only}} 
& \multirow{4}{*}{\textbf{Corp-HUS (Fr)}} 
& Our model (200) & 0.827 (0.004)* & 0.756 (0.009)* & 0.79 (0.006)* & 8 \\
& & Our model (300) & 0.834 (0.016)* & 0.782 (0.008)* & 0.807 (0.005) & 8 \\
& & Our model (300) + Frame RE & \textbf{0.852 (0.02)*} & 0.793 (0.012)* & \textbf{0.821 (0.015)} & 12 \\
& & Yang et al. 2021\cite{yang2021clinicalrelationextractionusing}\dag & 0.786 (0.02)  & \textbf{0.834 (0.007)} & 0.809 (0.013) & 108 \\
\cline{2-7}
& \multirow{7}{*}{\textbf{N2c2 (En)}} 
& Our model (300) Reduced Corpus & 0.842 (0.023) & 0.782 (0.045) & 0.81 (0.026) & - \\
& & Our model (200) & 0.960 & 0.942 & 0.952 & 18 \\
& & Our model (300) & 0.955 & \textbf{0.955} & 0.955 & 18 \\
& & Yang et al. 2021\cite{yang2021clinicalrelationextractionusing} \dag & 0.955 & 0.843 & 0.949 & 413 \\
\cline{3-7}
& & Xu J et al. 2017 \cite{Xu2017UTH}\ddag & \textbf{0.972} & \textbf{0.955} & \textbf{0.963} & - \\
& & Chapman AB et al. 2018\cite{pmlr-v90-chapman18a}\ddag & 0.957 & 0.949 & 0.953 & - \\
& & Christopoulou et al. 2020\cite{christopoulou2020adverse,ju2020ensemble}\ddag & 0.946 & 0.948 & 0.947 & - \\
\cline{1-7}
\multirow{4}{*}{\rotatebox{90}{End-to-end}} 
& \textbf{Corp-HUS (Fr)} & Our model (300) & 0.699 & 0.672 & 0.686 & 78 \\[4ex]
& \textbf{N2c2 (En)} & Our model (300) & 0.880 & 0.770 & 0.821 & 83 \\[2ex]

\cline{1-7}
\end{tabular}
         }
\end{table*}

\subsection*{Error Analysis}
In the task of RE, our model exhibited lower recall compared to baseline models. This limitation was primarily attributed to the handling of modifiers that apply to multiple drugs. For instance, a date indicating the start of multiple prescriptions often extends beyond the boundaries of the sliding window, leading to missed relationships.
Adding same-frame relations between the different attributes of a drug was found to slightly improve results, particularly in cases involving multi-frame prescriptions.  This approach enhances the model’s ability to capture complex relationships and dependencies within drug attributes, resulting in more accurate extraction and representation. However, the limited presence of multi-frame prescriptions (less than 5\%) in the test set, make it difficult to fully evaluate the impact of such representation. Future work could benefit from larger datasets containing more examples of multi-frame scenarios to better quantify these improvements.

\subsection*{Carbon impact}
The selection of optimal hyperparameters for the proposed model required 6,110 minutes of computation time, consuming an estimated 57.39 kWh of energy and resulting in a carbon footprint of approximately 2.94~kg~$CO_2$, as calculated using the methodology outlined in \cite{Green_Algorithms}. This computation was performed using our specific infrastructure. It is important to note that the reported carbon impact only accounts for hyperparameter optimization. If the time spent on additional testing and implementation were included, the total computation time and carbon footprint would be approximately three times higher.

\section*{Discussion}
\subsection*{Annotation of French Corpus}
The F1 score for inter-annotator agreement underscores the inherent challenges associated of achieving reliable annotation for clinical text. This score shows the complexity of the automated RE task. The variability in human annotations highlights the importance of rigorous annotation guidelines and the benefit of machine-assisted pre-annotation, which offers room for improvements in consistency and efficiency\cite{NEVEOL2011310}. 

\subsection*{Interpretation of Results}
We demonstrate that transformer-based architectures can be effectively used for RE on French corpora and, more broadly, on real-world healthcare data. Despite the complexity of the task for human annotators, transformer-based models can perform remarkably well.

One key strength of our work, lies in the diversity of the French EHRs text, which extends beyond hospital discharge reports, unlike the n2c2 data. Our method appears robust to both context and language, demonstrating good performance on clinical data and challenge datasets in both French and English.
The model's lower recall, particularly in scenarios involving modifiers that span multiple drugs, suggests room for improvement. Future research could explore how to better capture long-range dependencies\cite{distanceRE}.

The limited size of our dataset constrains its performance compared to the n2c2 dataset. This appears to be primarily due to the limited annotated data. Performance seems adequate when using reduced subset of the n2c2 dataset. The micro-average F1 score demonstrates this, with 0.807 for our dataset versus 0.810 for the restricted n2c2 dataset.

When examining a restricted n2c2 dataset (Table \ref{tab:supp_RE_en}) subset, notable variations emerge across different relation types. The Duration-Drug relation showed an F1 score of 0.593 (0.11), in contrast to 0.781 on the French corpus (Table \ref{tab:supp_RE_french}). The Frequency-Drug relation showed more pronounced differences, with an F1 score of 0.845 (0.05) on the n2c2 subset compared to 91.3 on the French corpus.

 The Dosage/Strength relation comparison proved more nuanced due to differing annotation schemas. While the n2c2 dataset initially separates Dosage and Strength, our schema combines them into a single category.

For the Route-Drug relation, the results were more consistent, with an F1 score of 0.85 (0.045) on the n2c2 subset and 0.81 on the French corpus. However, the comparison of the Dosage/Strength relation was less straightforward. While the n2c2 dataset separates Dosage and Strength, our schema combines them into a single category. When aligning the n2c2 subset with our approach by merging these categories (Table \ref{tab:supp_RE_english_align}), the F1 score reached 0.891 (0.024), compared to 0.898 on the French corpus. These findings highlight the interesting influence of annotation schema design on performance. The variations across different relation types underscore the complexity of medical text annotation and the importance of testing algorithms in languages other than English.

The end-to-end performance is limited by the capacity of the NER component, indicating room for improvement in this initial task (see Tables \ref{tab:supp_EE_french} and \ref{tab:supp_EE_english}). 

\subsection*{Comparison With Previous Work}

The top three methods in the n2c2 challenge used complex algorithms for RE to achieve high performance, with F1 scores of 0.963, 0.953, and 0.947. The first team \cite{Xu2017UTH} adopted a joint NER-RE method, enhanced by external resources (e.g. MedDRA dictionaries) and syntactic features like POS tagging. They classify potential entity pairs individually. This aligns conceptually with transformer-based tagging methods \cite{frustratingly}.

The second team \cite{pmlr-v90-chapman18a} used a feature-rich ensemble of Random Forest classifiers, using entity types, word forms, n-grams, and syntactic data for RE across three-sentence windows.
The third team \cite{christopoulou2020adverse, ju2020ensemble} combined models using multiple embeddings.

Our approach achieves comparable performance to these state-of-the-art techniques while being simpler, reproducible, and computationally efficient. Our model’s F1 score of 0.955 on the n2c2 dataset matches or surpasses prior transformer-based methods \cite{Mahendran2021-dj,Wei2020-ab}. Additionally, our F1 score of 0.821 on the French corpus (using frame-based representation) sets a strong benchmark for underexplored French applications. For NER (see appendix), the results are comparable to those reported in previous studies on French clinical data~\cite{jouffroy}.

Our approach successfully addressed several methodological challenges in drugs RE. By eliminating extensive preprocessing requirements and using a sliding window approach, we simplified the data preparation pipeline. The ability to generalize across different transformer models (ClinicalBERT, BioBERT, CamemBERT-BIO) further highlights the robustness of our method.

\subsection*{Key Contributions}
\subsubsection*{Computational time}
This study introduces an innovative framework for drugs RE from EHRs, offering both technical advancements and practical benefits. 
By deploying a computational setup within a hospital, we provided a real-world example of practical implementation. All experiments were run in a controlled environment to ensure consistency and reproducibility. The significant reduction in training time—10 times faster on the French corpus and 23 times faster on the n2c2 dataset—demonstrates the potential for more sustainable AI approaches in healthcare settings. This efficiency is particularly crucial in low-resource environments and for institutions with limited computational infrastructure.
Generative transformer models were not tested as they could not be used in our environment due to their larger VRAM requirements.

The carbon impact analysis provides transparency about the environmental costs of model development. The hyperparameter optimization alone consumed approximately 57.39 kWh and generated 2.94 kg of CO2 emissions. While this might seem modest, it underscores the need for continued research into computationally efficient machine learning techniques.

\subsubsection*{Representation of treatment regimen as frames}

Several challenges have been proposed for extracting drugs from EHRs, each focusing on a specific aspect of the task. In 2018, the MADE challenge \cite{made_challenge} and the n2c2 (Natural Language Processing for Clinical Data) challenge\cite{Henry2020-wp} concentrated on extracting drugs and their attributes. In 2022 \cite{n2c2_2022}, n2c2 addressed medication modifications such as stopping, starting, increasing, or decreasing dosages and shifted its focus to the extraction of context. However, no challenge has yet addressed the complete task of extracting drugs, their attributes, and their temporal context. 
Our annotation schema differs from the n2c2 dataset in several key aspects, although certain relation types are shared between the two datasets. 
A particular improvement was the introduction of the "frame" concept, which provides a more nuanced representation of drugs relationships. This approach captures drug-specific attributes and their relationships within a contextualized framework, enabling a more nuanced representation of drugs relationships. The frame-based representation proved particularly effective, improving the F1 score from 0.807 to 0.821 on the French corpus.
Augmenting annotated data  by adding relations between drug attributes could facilitate the extraction of multi-frame information from free text and potentially improve performance.
The limited presence of multi-frame prescriptions in the test set restricts a comprehensive evaluation of the frame-based approach. Further analysis is necessary to fully explore the potential of this methodology, which was used for cancer phenotype extraction~\cite{savova2017deepphe}. Expanding the dataset to include 
more 
complex medication regimens, such as those of cancer patients, would provide a more robust validation. These patients often undergo multiple chemotherapy line changes, offering valuable examples of complex, multi-frame prescriptions.

\subsubsection*{Cross-language evaluation}
Despite the limitations imposed by the small amount of manually labeled data, which affects the performance of RE, especially in end-to-end.
Our research made significant strides in clinical text in English and French, demonstrating the model's adaptability by achieving comparable performance on both English and French corpora. This is particularly important given the predominance of English-language research in clinical natural language processing. By validating the proposed architecture on a French corpus, we address the scarcity of research on clinical corpora in languages other than English and highlight its potential for broader applicability across diverse linguistic contexts.

\subsection*{Implications for Clinical Informatics}
This research has significant implications for clinical informatics, offering a more precise and computationally efficient method for extracting treatment regimens from EHRs. The improved relationship extraction could support more accurate medication tracking, potentially enhancing patient safety and therapeutic management.
  
\subsection*{Future Research}

Applying this algorithm to longitudinal data could enable the extraction of detailed medication timelines for individual patients. However, the development of reconstruction algorithms is still required to accurately assemble and represent these timelines.

\section*{Conclusion}
This study introduces a novel architecture for RE from clinical data, validated on the n2c2 dataset and applied to French clinical data, demonstrating its robustness across languages and contexts. It combines classical attribute extraction like frequency, dosage with contextual attribute extraction like starting date, negation and more. We define a new scheme of attribute annotation, enhancing the representation of drugs, their attributes, and their context.

  By presenting a novel, computationally efficient approach to drugs RE, our study contributes to the advancement of natural language processing in clinical settings. The proposed framework not only improves performance but also addresses critical challenges of computational complexity and data representation. 


\newpage
\section*{Supplementary material}

\renewcommand{\thetable}{S\arabic{table}}
\setcounter{table}{0}

\begin{table*}[h]
\centering
\caption{Relation extraction on French Corpus, Corp-HUS}
\label{tab:supp_RE_french}
\begin{tabular}{lcccr}

     \textbf{Relation}              & \textbf{Prec.}         & \textbf{Rec.}          & \textbf{F1}            & \textbf{Support} \\
                \hline
Start           & 0.759 (0.069) & 0.657 (0.02)  & 0.703 (0.039) & 121                          \\
Stop            & 0.716 (0.038) & 0.700 (0.058)   & 0.706 (0.030)  & 70                           \\
Ongoing         & 0.829 (0.025) & 0.767 (0.075) & 0.796 (0.050)  & 123                          \\
Increase        & 0.441 (0.110)  & 0.560 (0.152)  & 0.487 (0.116) & 10                           \\
Decrease        & 0.573 (0.292) & 0.300 (0.235)   & 0.353 (0.214) & 10                           \\
Negation        & 0.598 (0.048) & 0.667 (0.059) & 0.630 (0.044)  & 12                           \\
Contraindicated & 0.413 (0.368) & 0.200 (0.000)       & 0.229 (0.093) & 5                            \\
Hypothetical    & 0.690 (0.223)  & 0.511 (0.099) & 0.572 (0.095) & 9                            \\
Coref           & 0.810 (0.041)  & 0.565 (0.038) & 0.664 (0.019) & 34                           \\
Discontinue     & 0 (0)         & 0 (0)         & 0 (0)         & 1                            \\
\textbf{Overall (micro)} & \textbf{0.852 (0.02)}  & \textbf{0.793 (0.012)} & \textbf{0.821 (0.015)}  & \textbf{395}   \\     
 \hline
\end{tabular}
\end{table*}

\begin{table*}[h]
\centering
\caption{Relation extraction on Corp-HUS, with categories align to n2c2 categories}
\label{tab:supp_RE_english_align}
\begin{tabular}{lccc}

     \textbf{Relation}              & \textbf{Prec.}         & \textbf{Rec.}          & \textbf{F1}            \\
                \hline

        Dosage → Drug &  0.971 & 0.971&  0.971 \\
		Duration → Drug  & 0.800  &0.762 & 0.781 \\
        		Frequency → Drug 	&0.930  &0.896  &0.912   \\     
        Route → Drug &  0.814  &0.800 & 0.807\\
 \hline
\end{tabular}
\end{table*}

\begin{table}[h]
\centering
\caption{Relation extraction for English Corpus}
\label{tab:supp_RE_en}
\begin{tabular}{lccc|ccc}

    & \multicolumn{3}{c|}{\textbf{Entire Dataset}}      & \multicolumn{3}{c}{\textbf{Restricted Dataset}}     \\ 
         \textbf{Relation}                  & \textbf{Prec.} & \textbf{Rec.} & \textbf{F1}     & \textbf{Prec.} & \textbf{Rec.} & \textbf{F1}     \\ \hline
Strength → Drug         & 0.988          & 0.987         & 0.988          & 0.913 (0.016)  & 0.900 (0.047) & 0.906 (0.027)  \\
Dosage → Drug           & 0.983          & 0.979         & 0.981          & 0.888 (0.014)  & 0.868 (0.025) & 0.877 (0.009)  \\
Duration → Drug         & 0.928          & 0.880         & 0.904          & 0.798 (0.057)  & 0.493 (0.159) & 0.593 (0.112)  \\
Frequency → Drug        & 0.986          & 0.971         & 0.978          & 0.926 (0.028)  & 0.780 (0.082) & 0.845 (0.050)  \\
Form → Drug             & 0.993          & 0.988         & 0.991          & 0.973 (0.007)  & 0.906 (0.030) & 0.938 (0.015)  \\
Route → Drug            & 0.980          & 0.985         & 0.983          & 0.899 (0.018)  & 0.805 (0.085) & 0.847 (0.045)  \\
Reason → Drug           & 0.838          & 0.772         & 0.804          & 0.490 (0.021)  & 0.519 (0.078) & 0.501 (0.034)  \\
ADE → Drug              & 0.850          & 0.735         & 0.789          & 0.575 (0.050)  & 0.354 (0.065) & 0.435 (0.059)  \\ 
\textbf{Overall (micro)} & \textbf{0.962} & \textbf{0.942} & \textbf{0.952} & \textbf{0.842 (0.023)} & \textbf{0.782 (0.045)} & \textbf{0.810 (0.026)} \\ 
\hline
\end{tabular}
\end{table}

\begin{table*}
\centering
\caption{Performance Metrics for Entities and Relations, End-to-end on French Corpus}
\label{tab:supp_EE_french}
\begin{tabular}{lcccccc}
                   & \multicolumn{3}{c}{Strict} & \multicolumn{3}{c}{Lenient}  \\ 
\cmidrule(l){2-4}\cmidrule(l){5-7}
                   & Prec. & Rec.  & F1         & Prec. & Rec.  & F1           \\ 
\midrule
\textbf{Entities}  &       &       &            &       &       &              \\
Drug                & 0.947 & 0.961 & 0.954      & 0.964 & 0.976 & 0.970        \\
Drug Classe            & 0.779 & 0.853 & 0.814      & 0.789 & 0.863 & 0.824        \\
Condition          & 0.737 & 0.778 & 0.757      & 0.895 & 0.944 & 0.919        \\
Context           & 0.682 & 0.674 & 0.678      & 0.774 & 0.756 & 0.765        \\
Date               & 0.941 & 0.969 & 0.955      & 0.970 & 0.979 & 0.974        \\
Relative Date     & 0.674 & 0.818 & 0.739      & 0.751 & 0.912 & 0.824        \\
Dosage             & 0.954 & 0.971 & 0.963      & 0.960 & 0.977 & 0.968        \\
Duration              & 0.692 & 0.857 & 0.766      & 0.731 & 0.905 & 0.809        \\
Frequency               & 0.832 & 0.853 & 0.842      & 0.963 & 0.953 & 0.958        \\
Route              & 0.836 & 0.889 & 0.862      & 0.881 & 0.937 & 0.908        \\ 
\cmidrule(l){2-7}
Overall (micro)    & 0.860 & 0.898 & 0.878      & 0.909 & 0.931 & 0.920        \\ 
\hline\hline
       &       &            &       &       &              \\
\textbf{Relations} &       &       &            &       &       &              \\
Refer to          & 0.786 & 0.786 & 0.786      & 0.850 & 0.839 & 0.844        \\
Start              & 0.493 & 0.567 & 0.527      & 0.537 & 0.608 & 0.570        \\
Stop               & 0.600 & 0.581 & 0.590      & 0.650 & 0.629 & 0.639        \\ 
Ongoing         & 0.346 & 0.293 & 0.317      & 0.346 & 0.293 & 0.317        \\
Duration prescription    & 0.542 & 0.619 & 0.578      & 0.583 & 0.667 & 0.622        \\
Increase       & 0.455 & 0.500 & 0.476      & 0.455 & 0.500 & 0.476        \\
Decrease         & 0.444 & 0.400 & 0.421      & 0.444 & 0.400 & 0.421        \\
Administration Time & - & - & -      & - & - & -        \\
Negation           & 0.250 & 0.167 & 0.200      & 0.500 & 0.333 & 0.400        \\
Contraindicated         & 0.000 & 0.000 & 0.000      & 0.000 & 0.000 & 0.000        \\
Hypothetical       & 1.000 & 1.000 & 1.000      & 1.000 & 1.000 & 1.000        \\
Experiencer Time & - & - & -      & - & - & -        \\
Coref             & 0.708 & 0.500 & 0.586      & 0.750 & 0.529 & 0.621        \\
Discontinues         & 0.000 & 0.000 & 0.000      & 0.000 & 0.000 & 0.000        \\
\cmidrule(l){2-7}
Overall (micro)    & 0.649 & 0.630 & 0.639      & 0.699 & 0.673 & 0.686 \\       \hline
\end{tabular}
\end{table*}

\begin{table*}
\centering
\caption{Performance Metrics for Entities and Relations, End-to-end on English Corpus}
\label{tab:supp_EE_english}
\begin{tabular}{lllllll}
                 &        & Strict &        &        & Lenient &         \\ 
\cmidrule(l){2-4}\cmidrule(l){5-7}
                 & Prec.  & Rec.   & F1     & Prec.  & Rec.    & F1      \\ 
\hline
\textbf{Entities}         &        &        &        &        &         &         \\
Drug             & 0.913  & 0.931  & 0.922  & 0.948  & 0.962   & 0.955   \\
Strength         & 0.946  & 0.953  & 0.949  & 0.980  & 0.985   & 0.982   \\
Duration         & 0.719  & 0.738  & 0.729  & 0.858  & 0.878   & 0.868   \\
Route            & 0.929  & 0.923  & 0.926  & 0.950  & 0.941   & 0.945   \\
Form             & 0.902  & 0.896  & 0.899  & 0.963  & 0.941   & 0.952   \\
Ade              & 0.615  & 0.013  & 0.025  & 0.692  & 0.014   & 0.028   \\
Dosage           & 0.883  & 0.921  & 0.902  & 0.920  & 0.958   & 0.939   \\
Reason           & 0.594  & 0.661  & 0.626  & 0.670  & 0.739   & 0.703   \\
Frequency        & 0.822  & 0.850  & 0.836  & 0.961  & 0.983   & 0.972   \\ 
\cmidrule(l){2-7}

Overall (micro)  & 0.874  & 0.877  & 0.876  & 0.928  & 0.926   & 0.927   \\ 
\hline\hline
                 &        &        &        &        &         &         \\
\textbf{Relations }       &        &        &        &        &         &         \\
Strength → Drug  & 0.858  & 0.916  & 0.886  & 0.958  & 0.972   & 0.965   \\
Dosage → Drug    & 0.783  & 0.879  & 0.828  & 0.867  & 0.937   & 0.901   \\
Duration → Drug  & 0.540  & 0.681  & 0.602  & 0.735  & 0.838   & 0.783   \\
Frequency → Drug & 0.708  & 0.809  & 0.755  & 0.929  & 0.965   & 0.946   \\
Form → Drug      & 0.822  & 0.869  & 0.845  & 0.943  & 0.932   & 0.937   \\
Route → Drug     & 0.840  & 0.886  & 0.863  & 0.905  & 0.926   & 0.916   \\
Reason → Drug    & 0.406  & 0.548  & 0.467  & 0.497  & 0.633   & 0.557   \\
ADE → Drug       & 0.500  & 0.011  & 0.021  & 0.563  & 0.012   & 0.024   \\ 
\cmidrule(l){2-7}
 
Overall (micro)  & 0.726  & 0.794  & 0.758  & 0.844  & 0.871   & 0.857  \\
 \hline
\end{tabular}
\end{table*}


\begin{thebibliography}{10}

\bibitem{demner2009can}
Dina Demner-Fushman, Wendy~W Chapman, and Clement~J McDonald.
\newblock What can natural language processing do for clinical decision support?
\newblock {\em Journal of biomedical informatics}, 42(5):760--772, 2009.

\bibitem{escudie2017novel}
Jean-Baptiste Escudi{\'e}, Bastien Rance, Georgia Malamut, Sherine Khater, Anita Burgun, Christophe Cellier, and Anne-Sophie Jannot.
\newblock A novel data-driven workflow combining literature and electronic health records to estimate comorbidities burden for a specific disease: a case study on autoimmune comorbidities in patients with celiac disease.
\newblock {\em BMC medical informatics and decision making}, 17:1--10, 2017.

\bibitem{uzuner2010extracting}
{\"O}zlem Uzuner, Imre Solti, and Eithon Cadag.
\newblock Extracting medication information from clinical text.
\newblock {\em Journal of the American Medical Informatics Association}, 17(5):514--518, 2010.

\bibitem{MODI2024104603}
Salisu Modi, Khairul~Azhar Kasmiran, Nurfadhlina {Mohd Sharef}, and Mohd~Yunus Sharum.
\newblock Extracting adverse drug events from clinical notes: A systematic review of approaches used.
\newblock {\em Journal of Biomedical Informatics}, 151:104603, 2024.

\bibitem{FRAILENAVARRO2023105122}
David {Fraile Navarro}, Kiran Ijaz, Dana Rezazadegan, Hania Rahimi-Ardabili, Mark Dras, Enrico Coiera, and Shlomo Berkovsky.
\newblock Clinical named entity recognition and relation extraction using natural language processing of medical free text: A systematic review.
\newblock {\em International Journal of Medical Informatics}, 177:105122, 2023.
\newblock Review des articles qui s'interese au NER et RE. Avec les différetnes approches. Parle de l'importance de tester les algos en pratique, sur d'autres langues.

\bibitem{Landolsi2023}
Mohamed~Yassine Landolsi, Lobna Hlaoua, and Lotfi Ben Romdhane.
\newblock Information extraction from electronic medical documents: state of the art and future research directions.
\newblock {\em Knowledge and Information Systems}, 65(2):463--516, Feb 2023.

\bibitem{BERT}
Jacob Devlin, Ming-Wei Chang, Kenton Lee, and Kristina Toutanova.
\newblock {BERT}: Pre-training of deep bidirectional transformers for language understanding.
\newblock In Jill Burstein, Christy Doran, and Thamar Solorio, editors, {\em Proceedings of the 2019 Conference of the North {A}merican Chapter of the Association for Computational Linguistics: Human Language Technologies, Volume 1 (Long and Short Papers)}, pages 4171--4186, Minneapolis, Minnesota, June 2019. Association for Computational Linguistics.

\bibitem{GPT}
Alec Radford and Karthik Narasimhan.
\newblock Improving language understanding by generative pre-training.
\newblock 2018.

\bibitem{gema-etal-2024-edinburgh}
Aryo Gema, Giwon Hong, Pasquale Minervini, Luke Daines, and Beatrice Alex.
\newblock {E}dinburgh clinical {NLP} at {S}em{E}val-2024 task 2: Fine-tune your model unless you have access to {GPT}-4.
\newblock In Atul~Kr. Ojha, A.~Seza Do{\u{g}}ru{\"o}z, Harish Tayyar~Madabushi, Giovanni Da~San~Martino, Sara Rosenthal, and Aiala Ros{\'a}, editors, {\em Proceedings of the 18th International Workshop on Semantic Evaluation (SemEval-2024)}, pages 1894--1904, Mexico City, Mexico, June 2024. Association for Computational Linguistics.

\bibitem{Carrell2017}
David~S Carrell, Robert~E Schoen, Daniel~A Leffler, Michele Morris, Sherri Rose, Andrew Baer, Seth~D Crockett, Rebecca~A Gourevitch, Katie~M Dean, and Ateev Mehrotra.
\newblock Challenges in adapting existing clinical natural language processing systems to multiple, diverse health care settings.
\newblock {\em J Am Med Inform Assoc}, 24(5):986--991, September 2017.

\bibitem{miller-etal-2021-domain}
Timothy Miller, Egoitz Laparra, and Steven Bethard.
\newblock Domain adaptation in practice: Lessons from a real-world information extraction pipeline.
\newblock In Eyal Ben-David, Shay Cohen, Ryan McDonald, Barbara Plank, Roi Reichart, Guy Rotman, and Yftah Ziser, editors, {\em Proceedings of the Second Workshop on Domain Adaptation for NLP}, pages 105--110, Kyiv, Ukraine, April 2021. Association for Computational Linguistics.

\bibitem{baker-etal-1998-berkeley-framenet}
Collin~F. Baker, Charles~J. Fillmore, and John~B. Lowe.
\newblock The {B}erkeley {F}rame{N}et project.
\newblock In {\em 36th Annual Meeting of the Association for Computational Linguistics and 17th International Conference on Computational Linguistics, Volume 1}, pages 86--90, Montreal, Quebec, Canada, August 1998. Association for Computational Linguistics.

\bibitem{Henry2020-wp}
Sam Henry, Kevin Buchan, Michele Filannino, Amber Stubbs, and Ozlem Uzuner.
\newblock 2018 n2c2 shared task on adverse drug events and medication extraction in electronic health records.
\newblock {\em J Am Med Inform Assoc}, 27(1):3--12, January 2020.

\bibitem{survey}
Kartik Detroja, C.K. Bhensdadia, and Brijesh~S. Bhatt.
\newblock A survey on relation extraction.
\newblock {\em Intelligent Systems with Applications}, 19:200244, 2023.

\bibitem{comprehensive_survey}
Xiaoyan Zhao, Yang Deng, Min Yang, Lingzhi Wang, Rui Zhang, Hong Cheng, Wai Lam, Ying Shen, and Ruifeng Xu.
\newblock A comprehensive survey on relation extraction: Recent advances and new frontiers.
\newblock {\em ACM Comput. Surv.}, 56(11), July 2024.

\bibitem{frustratingly}
Zexuan Zhong and Danqi Chen.
\newblock A frustratingly easy approach for entity and relation extraction.
\newblock In Kristina Toutanova, Anna Rumshisky, Luke Zettlemoyer, Dilek Hakkani-Tur, Iz~Beltagy, Steven Bethard, Ryan Cotterell, Tanmoy Chakraborty, and Yichao Zhou, editors, {\em Proceedings of the 2021 Conference of the North American Chapter of the Association for Computational Linguistics: Human Language Technologies}, pages 50--61, Online, June 2021. Association for Computational Linguistics.

\bibitem{zheng-etal-2017-joint}
Suncong Zheng, Feng Wang, Hongyun Bao, Yuexing Hao, Peng Zhou, and Bo~Xu.
\newblock Joint extraction of entities and relations based on a novel tagging scheme.
\newblock In Regina Barzilay and Min-Yen Kan, editors, {\em Proceedings of the 55th Annual Meeting of the Association for Computational Linguistics (Volume 1: Long Papers)}, pages 1227--1236, Vancouver, Canada, July 2017. Association for Computational Linguistics.

\bibitem{mtumbuka-schockaert-2024-entity}
Frank~Martin Mtumbuka and Steven Schockaert.
\newblock Entity or relation embeddings? an analysis of encoding strategies for relation extraction.
\newblock In Yaser Al-Onaizan, Mohit Bansal, and Yun-Nung Chen, editors, {\em Findings of the Association for Computational Linguistics: EMNLP 2024}, pages 6003--6022, Miami, Florida, USA, November 2024. Association for Computational Linguistics.

\bibitem{Mahendran2021-dj}
Darshini Mahendran and Bridget~T McInnes.
\newblock Extracting adverse drug events from clinical notes.
\newblock {\em AMIA Jt Summits Transl Sci Proc}, 2021:420--429, May 2021.

\bibitem{Wei2020-ab}
Qiang Wei, Zongcheng Ji, Yuqi Si, Jingcheng Du, Jingqi Wang, Firat Tiryaki, Stephen Wu, Cui Tao, Kirk Roberts, and Hua Xu.
\newblock Relation extraction from clinical narratives using pre-trained language models.
\newblock {\em AMIA Annu Symp Proc}, 2019:1236--1245, March 2020.

\bibitem{yang2021clinicalrelationextractionusing}
Xi~Yang, Zehao Yu, Yi~Guo, Jiang Bian, and Yonghui Wu.
\newblock Clinical relation extraction using transformer-based models, 2021.

\bibitem{Zhang}
Zhenyu Zhang, Lin Shi, Yang Yuan, Huanyue Zhou, and Shoukun Xu.
\newblock Multi-level attention with 2d table-filling for joint entity-relation extraction.
\newblock {\em Information}, 15:407, 07 2024.
\newblock un peu la même approche mais dans la matrice de prédiction rajoute des lignes et colonne pour la relation.

\bibitem{clinicalbert}
Kexin Huang, Jaan Altosaar, and Rajesh Ranganath.
\newblock Clinicalbert: Modeling clinical notes and predicting hospital readmission.
\newblock {\em arXiv:1904.05342}, 2019.

\bibitem{biobert}
Jinhyuk Lee, Wonjin Yoon, Sungdong Kim, Donghyeon Kim, Sunkyu Kim, Chan~Ho So, and Jaewoo Kang.
\newblock Biobert: a pre-trained biomedical language representation model for biomedical text mining.
\newblock {\em Bioinformatics}, 36(4):1234--1240, 09 2019.

\bibitem{camembert_bio}
Rian Touchent and {\'E}ric de~la Clergerie.
\newblock {C}amem{BERT}-bio: Leveraging continual pre-training for cost-effective models on {F}rench biomedical data.
\newblock In Nicoletta Calzolari, Min-Yen Kan, Veronique Hoste, Alessandro Lenci, Sakriani Sakti, and Nianwen Xue, editors, {\em Proceedings of the 2024 Joint International Conference on Computational Linguistics, Language Resources and Evaluation (LREC-COLING 2024)}, pages 2692--2701, Torino, Italia, May 2024. ELRA and ICCL.

\bibitem{wajsburt}
Perceval Wajsb{\"u}rt.
\newblock {\em {Extraction and normalization of simple and structured entities in medical documents}}.
\newblock Theses, {Sorbonne Universit{\'e}}, December 2021.

\bibitem{tensorboard}
Mart\'{i}n Abadi, Ashish Agarwal, Paul Barham, Eugene Brevdo, Zhifeng Chen, Craig Citro, Greg~S. Corrado, Andy Davis, Jeffrey Dean, Matthieu Devin, Sanjay Ghemawat, Ian Goodfellow, Andrew Harp, Geoffrey Irving, Michael Isard, Yangqing Jia, Rafal Jozefowicz, Lukasz Kaiser, Manjunath Kudlur, Josh Levenberg, Dandelion Man\'{e}, Rajat Monga, Sherry Moore, Derek Murray, Chris Olah, Mike Schuster, Jonathon Shlens, Benoit Steiner, Ilya Sutskever, Kunal Talwar, Paul Tucker, Vincent Vanhoucke, Vijay Vasudevan, Fernanda Vi\'{e}gas, Oriol Vinyals, Pete Warden, Martin Wattenberg, Martin Wicke, Yuan Yu, and Xiaoqiang Zheng.
\newblock {TensorFlow}: Large-scale machine learning on heterogeneous systems, 2015.
\newblock Software available from tensorflow.org.

\bibitem{medkit}
Antoine Neuraz, Ghislain Vaillant, Camila Arias, Olivier Birot, Kim-Tam Huynh, Thibaut Fabacher, Alice Rogier, Nicolas Garcelon, Ivan Lerner, Bastien Rance, and Adrien Coulet.
\newblock Facilitating phenotyping from clinical texts: the medkit library.
\newblock {\em Bioinformatics}, 40(12):btae681, 11 2024.

\bibitem{pytorch}
Adam Paszke, Sam Gross, Francisco Massa, Adam Lerer, James Bradbury, Gregory Chanan, Trevor Killeen, Zeming Lin, Natalia Gimelshein, Luca Antiga, Alban Desmaison, Andreas Kopf, Edward Yang, Zachary DeVito, Martin Raison, Alykhan Tejani, Sasank Chilamkurthy, Benoit Steiner, Lu~Fang, Junjie Bai, and Soumith Chintala.
\newblock Pytorch: An imperative style, high-performance deep learning library.
\newblock In {\em Advances in Neural Information Processing Systems 32}, pages 8024--8035. Curran Associates, Inc., 2019.

\bibitem{transformers}
Thomas Wolf, Lysandre Debut, Victor Sanh, Julien Chaumond, Clement Delangue, Anthony Moi, Pierric Cistac, Tim Rault, Rémi Louf, Morgan Funtowicz, Joe Davison, Sam Shleifer, Patrick von Platen, Clara Ma, Yacine Jernite, Julien Plu, Canwen Xu, Teven~Le Scao, Sylvain Gugger, Mariama Drame, Quentin Lhoest, and Alexander~M. Rush.
\newblock Transformers: State-of-the-art natural language processing.
\newblock In {\em Proceedings of the 2020 Conference on Empirical Methods in Natural Language Processing: System Demonstrations}, pages 38--45, Online, October 2020. Association for Computational Linguistics.

\bibitem{pandas}
{W}es {M}c{K}inney.
\newblock {D}ata {S}tructures for {S}tatistical {C}omputing in {P}ython.
\newblock In {S}t\'efan van~der {W}alt and {J}arrod {M}illman, editors, {\em {P}roceedings of the 9th {P}ython in {S}cience {C}onference}, pages 56 -- 61, 2010.

\bibitem{numpy}
Charles~R. Harris, K.~Jarrod Millman, St{\'{e}}fan~J. van~der Walt, Ralf Gommers, Pauli Virtanen, David Cournapeau, Eric Wieser, Julian Taylor, Sebastian Berg, Nathaniel~J. Smith, Robert Kern, Matti Picus, Stephan Hoyer, Marten~H. van Kerkwijk, Matthew Brett, Allan Haldane, Jaime~Fern{\'{a}}ndez del R{\'{i}}o, Mark Wiebe, Pearu Peterson, Pierre G{\'{e}}rard-Marchant, Kevin Sheppard, Tyler Reddy, Warren Weckesser, Hameer Abbasi, Christoph Gohlke, and Travis~E. Oliphant.
\newblock Array programming with {NumPy}.
\newblock {\em Nature}, 585(7825):357--362, September 2020.

\bibitem{seqeval}
Hiroki Nakayama.
\newblock {seqeval}: A python framework for sequence labeling evaluation, 2018.
\newblock Software available from https://github.com/chakki-works/seqeval.

\bibitem{Green_Algorithms}
Lo{\"\i}c Lannelongue, Jason Grealey, and Michael Inouye.
\newblock Green algorithms: Quantifying the carbon footprint of computation.
\newblock {\em Adv Sci (Weinh)}, 8(12):2100707, May 2021.

\bibitem{Xu2017UTH}
Jun Xu, Hee-Jin Lee, Zongcheng Ji, Jingqi Wang, Qiang Wei, and Hua Xu.
\newblock Uth\_ccb system for adverse drug reaction extraction from drug labels at tac-adr 2017.
\newblock {\em Theory and Applications of Categories}, 2017.

\bibitem{pmlr-v90-chapman18a}
Alec~B. Chapman, Kelly~S. Peterson, Patrick~R. Alba, Scott~L. DuVall, and Olga~V. Patterson.
\newblock Hybrid system for adverse drug event detection.
\newblock In Feifan Liu, Abhyuday Jagannatha, and Hong Yu, editors, {\em Proceedings of the 1st International Workshop on Medication and Adverse Drug Event Detection}, volume~90 of {\em Proceedings of Machine Learning Research}, pages 16--24. PMLR, 04 May 2018.

\bibitem{christopoulou2020adverse}
Fenia Christopoulou, Thy~Thy Tran, Sunil~Kumar Sahu, Makoto Miwa, and Sophia Ananiadou.
\newblock Adverse drug events and medication relation extraction in electronic health records with ensemble deep learning methods.
\newblock {\em Journal of the American Medical Informatics Association}, 27(1):39--46, 2020.

\bibitem{ju2020ensemble}
Meizhi Ju, Nhung~TH Nguyen, Makoto Miwa, and Sophia Ananiadou.
\newblock An ensemble of neural models for nested adverse drug events and medication extraction with subwords.
\newblock {\em Journal of the American Medical Informatics Association}, 27(1):22--30, 2020.

\bibitem{NEVEOL2011310}
Aurélie Névéol, Rezarta {Islamaj Doğan}, and Zhiyong Lu.
\newblock Semi-automatic semantic annotation of pubmed queries: A study on quality, efficiency, satisfaction.
\newblock {\em Journal of Biomedical Informatics}, 44(2):310--318, 2011.

\bibitem{distanceRE}
Xiao Wei and Yongqi Chen.
\newblock Joint extraction of long-distance entity relation by aggregating local- and semantic-dependent features.
\newblock {\em Wireless Communications and Mobile Computing}, 2022(1):3763940, 2022.

\bibitem{jouffroy}
Jordan Jouffroy, Sarah~F Feldman, Ivan Lerner, Bastien Rance, Anita Burgun, and Antoine Neuraz.
\newblock Hybrid deep learning for medication-related information extraction from clinical texts in french: Medext algorithm development study.
\newblock {\em JMIR Med Inform}, 9(3):e17934, Mar 2021.

\bibitem{made_challenge}
Abhyuday Jagannatha, Feifan Liu, Weisong Liu, and Hong Yu.
\newblock Overview of the first natural language processing challenge for extracting medication, indication, and adverse drug events from electronic health record notes ({MADE} 1.0).
\newblock {\em Drug Saf}, 42(1):99--111, January 2019.

\bibitem{n2c2_2022}
Diwakar Mahajan, Jennifer~J. Liang, Ching-Huei Tsou, and Özlem Uzuner.
\newblock Overview of the 2022 n2c2 shared task on contextualized medication event extraction in clinical notes.
\newblock {\em Journal of Biomedical Informatics}, 144:104432, 2023.

\bibitem{savova2017deepphe}
Guergana~K Savova, Eugene Tseytlin, Sean Finan, Melissa Castine, Timothy Miller, Olga Medvedeva, David Harris, Harry Hochheiser, Chen Lin, Girish Chavan, et~al.
\newblock Deepphe: a natural language processing system for extracting cancer phenotypes from clinical records.
\newblock {\em Cancer research}, 77(21):e115--e118, 2017.

\end{thebibliography}
\end{document}